# Data-driven prognostics based on time-frequency analysis and symbolic recurrent neural network for fuel cells under dynamic load

Chu Wang [a, b, c, *], Manfeng Dou [b], Zhongliang Li [c, **], Rachid Outbib [c], Dongdong Zhao [b], Jian Zuo [c, d], Yuanlin Wang [b], Bin Liang [e], Peng Wang [a, f, ***]

[a] School of Electronic Information Engineering, Xi'an Technological University, Xi'an 710021, China

[b] School of Automation, Northwestern Polytechnical University, Xi'an 710072, China

[c] LIS Lab (UMR CNRS 7020), Aix-Marseille Université, 13397 Marseille, France

[d] GIPSA-lab (UMR CNRS 5216), Université Grenoble Alpes, 38402 Grenoble, France

[e] Department of Automation, Tsinghua University, Beijing 100084, China

[f] Office of Development and Planning, Xi'an Technological University, Xi'an 710021, China

[*] Corresponding author: Chu Wang (e-mail: chu.wang@etu.univ-amu.fr)

[**] Corresponding author: Zhongliang Li (e-mail: zhongliang.li@lis-lab.fr)

[***] Corresponding author: Peng Wang (e-mail: wang_peng@xatu.edu.cn)

**Abstract**

Data-centric prognostics is beneficial to improve the reliability and safety of proton exchange membrane fuel cell (PEMFC). For the prognostics of PEMFC operating under dynamic load, the challenges come from extracting degradation features, improving prediction accuracy, expanding the prognostics horizon, and reducing computational cost. To address these issues, this work proposes a data-driven PEMFC prognostics approach, in which Hilbert-Huang transform is used to extract health indicator in dynamic operating conditions and symbolic-based gated recurrent unit model is used to enhance the accuracy of life prediction. Comparing with other state-of-the-art methods, the proposed data-driven prognostics approach provides a competitive prognostics horizon with lower computational cost. The prognostics performance shows consistency and generalizability under different failure threshold settings.

    Dynamic load;

    Empirical mode decomposition;

    Time-frequency-energy spectrum;

    Symbolic representation gated recurrent unit;

    Remaining useful life;

# 1 Introduction

Fuel cells are carbon-emission avoiding energy conversion devices, which are considered one of the promising alternatives to fossil fuels [1]. Among different types, proton exchange membrane fuel cells (PEMFCs) dominate worldwide in terms of annual shipments and total power generation, and continue to rise [4]. Meanwhile, PEMFCs are adapted to diverse application requirements, such as transportation, stationary power generation and portable devices, which are attributed to advantages such as favourable start-up/operating temperature and high power density [2,3]. Nevertheless, the unsatisfactory durability and high cost still constrain the wider commercialization of PEMFCs. Besides, once PEMFCs are exposed to complex operating conditions, the performance degradation may accelerate and the lifetime will be even more inadequate. For instance, the average lifetime of fuel cell electric vehicles is about 2000 hours, while the performance will degrade by 30% after 5000 service hours [6]. Therefore, for PEMFCs operating under dynamic loads, the inadequate lifespan emerges as the core bottleneck [5].

Prognostics and health management (PHM) is a highly promising candidate technology for extending lifespan and enhancing durability. It assesses and predicts the evolving behavior of engineering equipment, systems and structures to enable the prediction of failures and the avoidance of accidents, and ultimately to achieve reliable, efficient, economical and safe operations [7,8]. Machine learning and deep learning are already successful in the fields of control and computer vision [9]. PHM techniques using similar data-centric approaches are also flourishing in reliability engineering and safety analysis [10]. As one of the keys to PHM, "Prognostics" tracks the health status of PEMFCs. Further, the remaining useful life (RUL) can be predicted to support condition-based maintenance (CBM) [3]. However, for PEMFCs operating under dynamic loads, deployment prognostics is fraught with challenges.

On the one hand, the measurements collected under dynamic mission profiles are difficult to use directly as health indicator (HI). If HI is not easily accessible or missing, then continuous PEMFC health monitoring becomes difficult, and even RUL predictions may not be achieved. Several scholars extract the inherent degradation parameters of PEMFC by mechanistic models and utilize filter methods to identify these parameters as HIs [11-13]. Li et al. deploy the linear parameter-varying (LPV) model in a series of sliding data segments and consider the reconstructed virtual steady-state stack voltage as HI [14]. Yue et al. present a polarization function-based model in [15] and segment the measured voltage to extract HI using a nonlinear regression method. Wang et al. utilize a degradation behaviour model to track the load dynamics and extract the equivalent internal resistance as HI by a sliding-window approach [16]. Hua et al. deploy a polarization test at the beginning of a lifetime and propose the relative power-loss rate as HI based on the observed power [17]. However, these methods have shortcomings such as high computational cost, requiring extra characterization tests, and hardly tracking transient dynamics. Zhang et al. propose a data-driven method based on Hilbert-Huang transform (HHT) to extract the HI of solid oxide fuel cell (SOFC) at constant operating condition [31]. In fact, HHT is a powerful tool for time-frequency analysis with the ability to extract inherent degradation trends from disturbances [31]. Meanwhile, the HHT is already being applied in the prognostics of complex systems/devices such as lithium-ion batteries [25-27], rolling element bearings [28], nuclear power plants [29], etc., and in the diagnostics of fuel cells [30]. Whereas, the use of HHT to extract health indicators of PEMFC under dynamic load conditions is not yet fully investigated.

On the other hand, the degradation behaviour of PEMFC is coupled with dynamic operations, which makes it difficult to predict both the inherent degradation trend and the RUL. Some scholars prefer to deploy short-term prognostics (single-step-ahead prediction) under dynamic mission profiles [18-20]. In practice, highly accurate single-step-ahead prediction is more meaningful for the real-time control and monitoring of fuel cells [21]. In contrast, long-term prognostics dedicated to RUL prediction requires extending the prognostics horizon (PH) to hundreds or even thousands of hours , in order to provide enough time for maintenance planning to avoid fatal failures [3]. In state-of-the-art works, extended Kalman filter (EKF) based methods [11,12], and echo state network (ESN) based methods [14,15,17] are used to predict the RUL under dynamic mission profiles. Most of these long-term prognostics methods are based on the multi-step-ahead prediction and set a fixed failure threshold (FT) for method evaluation. While setting only a single failure threshold, the performance of RUL prediction is difficult to be fully evaluated. In addition, whether filter-based or data-driven approaches are used, they usually require cautious model configuration, and it is necessary to reduce computational costs to improve real-time performance. Besides, thanks to the convincing time series processing capability, several LSTM framework-based RUL prediction methods are developed for different applactions [13,16,21,32-35]. Compared to LSTM, GRU effectively simplifies the hidden unit structure and reduces the number of parameters [36,37]. This facilitates the training/prediction efficiency of GRU and promises to improve the real-time prognostics performance. The GRU-based degradation prediction method is successfully applied in fields such as aero-propulsion system [38-40] and lithium-ion batteries [26,27,41-43]. As mentioned in our previous works [16,21,32,33], the performance of raw LSTM is not satisfactory in multi-step-ahead prediction in the application of fuel cell prognostics. Similarly, GRU also suffers from this issue, i.e., in the prediction phase, the prediction results will eventually converge to a horizontal line due to the lack of real-time observations to update the trained model, and no degradation trend prediction can be achieved.

In summary, although studies involving PEMFC durability are highly noted, long-term prognostics under dynamic loading profiles has not yet been fully addressed. The main challenges arise from the following two-fold questions: (i) How to extract the health indicator from dynamic mission profiles efficiently without disturbing PEMFC operation. (ii) How to extend the scale of prognostics horizon and to achieve stable prognostics performance under different failure thresholds. To cope with the above issues, a data-driven prognostics method based on time-frequency analysis and symbolic recurrent neural networks (RNNs) is proposed in this paper. Specifically, the long-term degradation component of the dynamic stack voltage is first extracted as HI using the HHT. The HI is compressed using adaptive Brownian bridge-based aggregation (ABBA) and represented as a reduced-dimensional symbolic sequence [22]. The trend of the represented symbolic sequence is predicted by gated recurrent unit (GRU)-based RNN, which in turn is used to reconstruct the HI trend and estimate the RUL. The proposed method is evaluated using dynamic load ageing experimental data from two different types of PEMFCs. In the HI extraction phase, the HHT-based method shows the lowest computational cost compared to other state-of-the-art methods. In the RUL prediction phase, the proposed data-driven approach provides competitive prognostics horizon and accuracy. Consistency and generalizability of prognostics performance are appropriately assessed using multiple failure thresholds.

## 2   Hilbert-Huang transform-based health indicator extraction

The PEMFC stack voltage is easy to collect and does not require the installation of additional test equipment. Numerous government agencies, research institutes and industries directly and/or indirectly use stack voltages when measuring PEMFC performance [48]. The HHT proposed by Huang et al. is adaptive, efficient, and considered as a suitable tool for processing non-linear and non-stationary signals [23]. While the PEMFC stack voltage coupled with dynamic load disturbances is exactly a typical class of non-linear and non-stationary characteristic signals. Therefore, HHT is utilized in this work to extract health indicators from dynamic voltages.

The HHT consists of two steps: First, the input signal is decomposed into a series of intrinsic mode functions (IMFs) and a residual using empirical mode decomposition (EMD). Second, the Hilbert spectrum of each IMF is obtained by deploying the Hilbert transform (HT). The combination of EMD and Hilbert Spectral Analysis (HSA) provides a time-frequency-energy analysis method [24]. Generally, EMD sifting is iteratively implemented and stopped under the condition that a monotonic residual is obtained. The residual reflects the trend characteristics in the original signal [31]. The residual contributes to revealing the physical characteristics of the signal as it is the low frequency or null component [24]. In this work, the dynamic voltage of the PEMFC is used as the input signal of EMD sifting. The proper residual from the input signal is identified by iteratively implementing EMD and designated as HI. Considering the instantaneous frequency (IF) is one of the key features for analyzing the signal based on natural conditions [24]. In this paper, an HI extraction process based on IF analysis is proposed as shown in Figure 1. Instead of pursuing a completely monotonic residual as the stop condition of the EMD sifting process, the method analyzes whether the IF of the residual is below a set threshold.

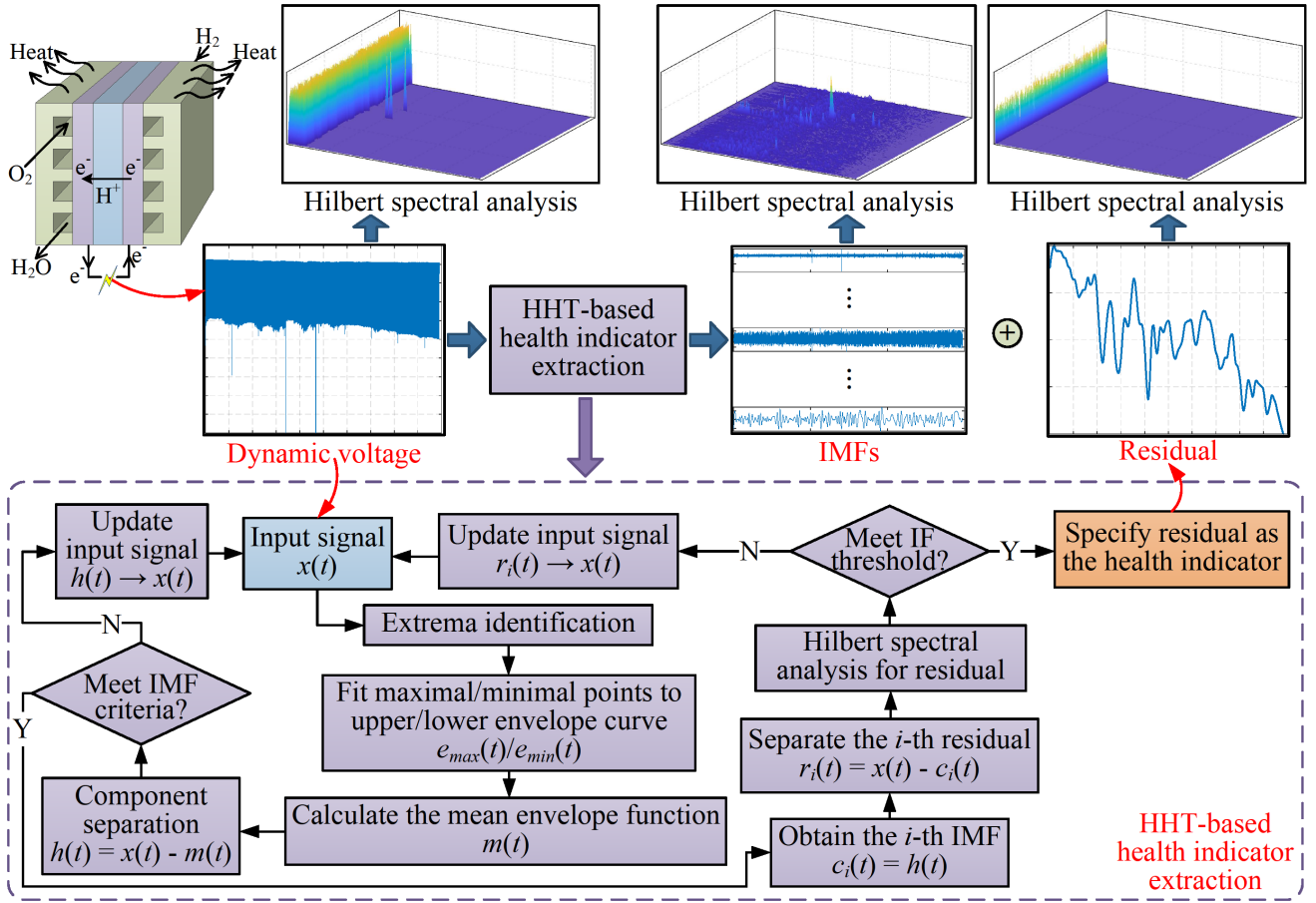

Figure 1. Hilbert-Huang transform-based health indicator extraction process.

As shown in Figure 1, the specific sifting process is explained as follows.

(a) Identify all extrema of the input signal $x(t)$. The upper envelope curve $e_{max}(t)$ and the lower envelope curve $e_{min}(t)$ are formed by fitting the maximal and minimal points through cubic spline curves, respectively. The mean envelope function $m_{1,1}(t)$ is calculated as below

$$m_{1,1}(t) = [e_{max}(t) + e_{min}(t)]/2 \qquad (1)$$

(b) Subtract $m_{1,1}(t)$ from $x(t)$ to obtain the first component $h_{1,1}(t)$,

$$h_{1,1}(t) = x(t) - m_{1,1}(t) \qquad (2)$$

Each IMF should satisfy the following sifting criteria [23]:

　　i.　The number of poles and zeros should not differ by more than one;

　　ii.　The mean of the upper and lower envelope curves should be zero.

$h_{1,1}(t)$ is defined as the first IMF $c_1(t)$ if it satisfies the criteria. Otherwise, repeat steps (a)-(b) until the requirements are satisfied at the $j$-th repetition,

$$h_{1,j}(t) = h_{1,j-1}(t) - m_{1,j}(t) \tag{3}$$

Then $h_{1,j}(t)$ is specified as $c_1(t)$.

(c) Subtract $c_1(t)$ from $x(t)$ to obtain the first residual $r_1(t)$,

$$r_1(t) = x(t) - c_1(t) \tag{4}$$

(d) The Hilbert transform is deployed only for the residual to obtain its IF, and the decomposition is stopped when the IF is less than a pre-set frequency threshold. Otherwise, the residual is used as the input signal and the above steps are repeated until the IF is satisfied at the $n$-th repetition. Then the final residual is $r_n(t)$ and is specified as HI,

$$r_n(t) = r_{n-1}(t) - c_n(t) \tag{5}$$

Up to this point, $x(t)$ can be expressed by the following equation,

$$x(t) = r_n(t) + \sum_{i=1}^{n} c_n(t) \tag{6}$$

**Remarks:** (1) The residual signal implies the long-term degradation trend characteristics of the PEMFC stack voltage. (2) The instantaneous frequency threshold is set based on both the time scale of PEMFC degradation and the time-frequency-energy characteristics of the stack voltage signal. (3) The basic formula for the Hilbert transform can be found in Appendix A.

## 3 Symbolic GRU-based lifetime prediction

The essential task of prognostics is to track the degradation trend of HI and predict the RUL. This work proposes an ABBA-based GRU (ABBA-GRU) model to tackle multi-step-ahead prediction involved in prognostics. The historical HI is used as input, while the output is the predicted trend. The core of the ABBA-GRU model includes three components: conversion, prediction, and reconstruction, as in Figure 2 (a) and Table 1. To improve prediction efficiency and accuracy, HI is normalized and re-scaled to between zero and one before being fed to ABBA-GRU.

Conversion, a. k. a. representation, is a data dimensionality reducing process that transforms a time series into a symbolic series. In the first step, the time series ($X$) is changed into the increments-tuple set ($D$) by compression. In other words, the core of Compression is to divide the n-dimensional input time series into m segments. In the second step, *D* is converted into the symbolic series (*A*) by digitization. Specifically, Digitization is to generate symbols by dividing clusters, and then to label *m* segments with these symbols. *A* is to be used as the input for the prediction phase.

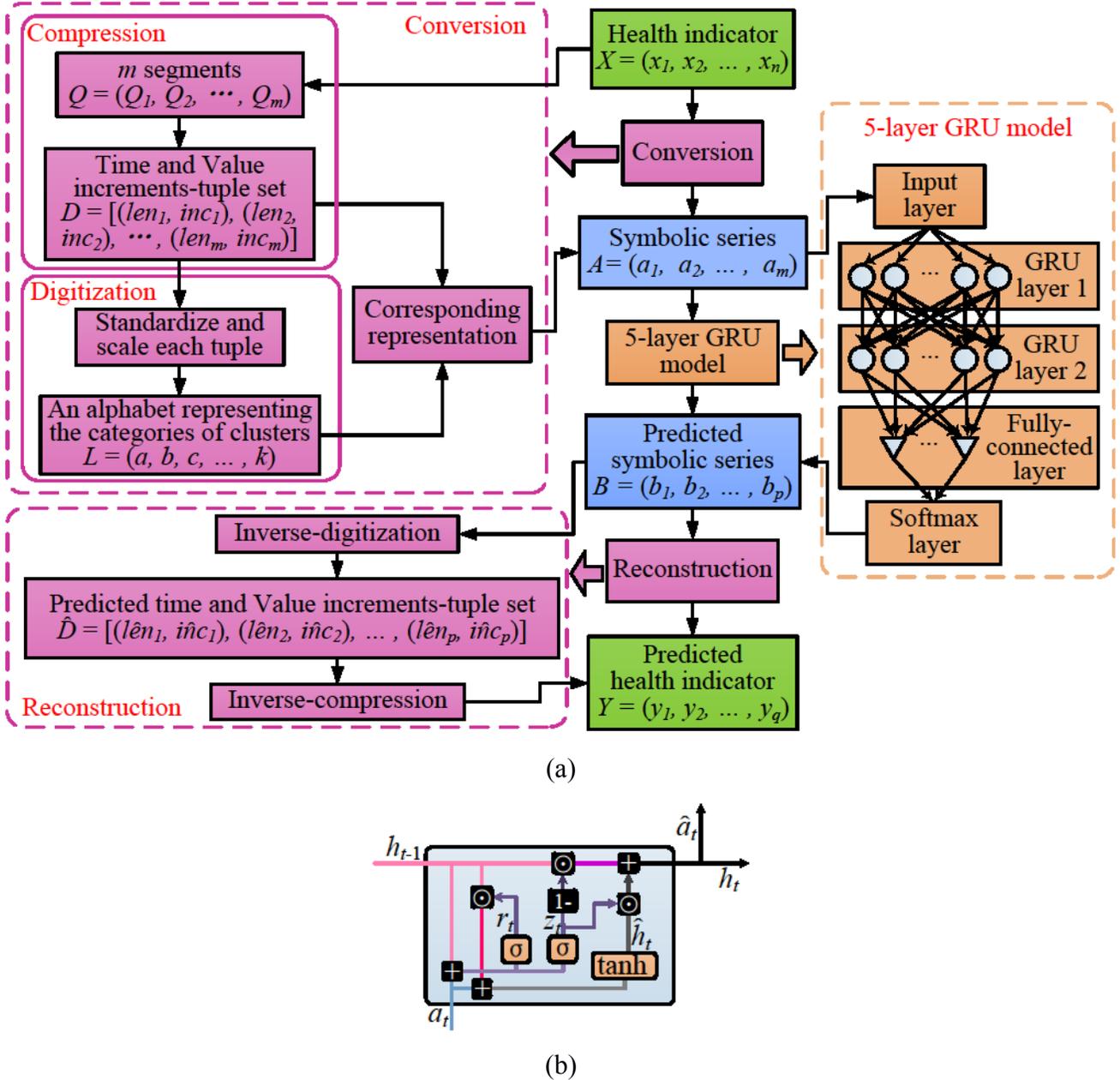

(a)

(b)

Figure 2. Symbolic GRU-based degradation trends prediction process, (a). ABBA-GRU structure; (b). GRU architecture.

A five-layer GRU model is used in the prediction phase, as in Figure 2 (a). Meanwhile, Figure 2 (b) shows the structure of a typical GRU hidden unit, about which a brief description can be found in Appendix B. For more details about GRU refer to the literature [36, 37].

Table 1 Degradation trend prediction steps and notation defination

| Step | | Data type | Notation |
|---|---|---|---|
| Input | | Time series | $X = (x_1, x_2, \cdots, x_n) \in \mathbb{R}^n$ |
| Representation | Compression | Time and numerical increments-tuple set | $D = \begin{bmatrix} (len_1, inc_1), (len_2, inc_2) \\ , \cdots, (len_m, inc_m) \end{bmatrix} \in \mathbb{R}^{2 \times m}$ |
| | Digitization | Alphabet set | $L = (a, b, c, \cdots, k)$ ($k$ types of clusters) |
| | | Symbolic series | $A = (a_1, a_2, \cdots, a_m) \in L^m$ |
| Prediction | | Symbolic series (predicted) | $B = (b_1, b_2, \cdots, b_p) \in L^p$ |
| Reconstruction | Inverse-digitization | Increments-tuple set (predicted) | $\widehat{D} = \begin{bmatrix} (l\hat{e}n_1, i\hat{n}c_1), (l\hat{e}n_2, i\hat{n}c_2) \\ , \cdots, (l\hat{e}n_p, i\hat{n}c_p) \end{bmatrix} \in \mathbb{R}^{2 \times p}$ |
| | Inverse-compression (Output) | Time series (predicted) | $Y = (y_1, y_2, \cdots, y_q) \in \mathbb{R}^q$ |

The reconstruction phase, including Inverse-digitization and Inverse-compression, can be considered as the inverse process of conversion. The predicted symbolic series ($B$) is changed into the predicted increments-tuple set ($\widehat{D}$) by Inverse-digitization. Subsequently, $\widehat{D}$ is changed into the predicted time series ($Y$) by Inverse-compression. More details about ABBA can be found in our previous work [16].

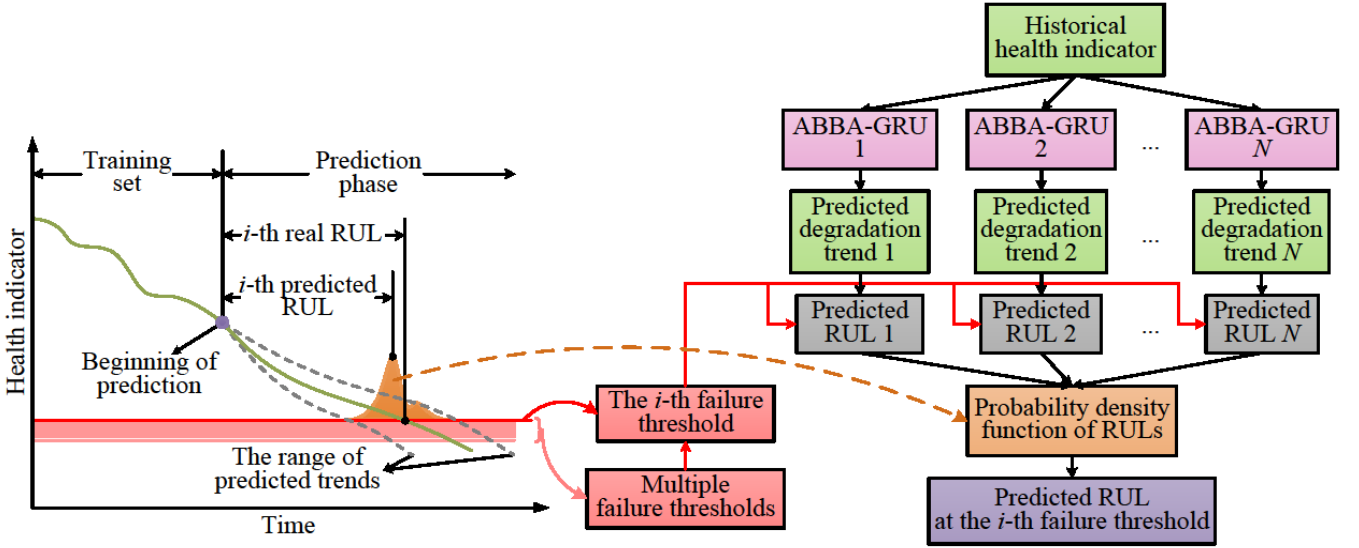

Figure 3. Remaining useful life prediction process and multiple failure thresholds evaluation.

This work sets multiple failure thresholds to evaluate the consistency and credibility of the prognostics approach. At the $i$-th failure threshold, as in Figure 3, multiple (here $N$) ABBA-GRU models are constructed by setting different initial weight matrices. Different initialization weight matrices can bring diversity to the GRU model and will be further used to improve the epistemic uncertainty of the model. Multiple predicted degradation trends are used to calculate the corresponding RULs respectively. Afterwards, the probability distribution function (PDF) of the RULs is generated. The RUL corresponding to the maximum of the PDF is considered the final prediction.

**Remarks**: (1) In this work, PDF is derived by kernel density estimation (KDE). In general, KDE is a nonparametric method for probability density estimation using kernel smoothing techniques, which is based on a limited sample of data to infer the overall population; (2) The vicinity of the maximum value of the PDF corresponds to the most active part of the output results from the prediction model. Specifying the estimated RUL corresponding to the maximum value of the PDF as the final result not only respects the experimental facts, but also facilitates the method reproducibility.

# 4  PEMFC ageing experiments under dynamic load

In this work, hardware experiments from two different PEMFCs are manipulated. They cover long-term aging data corresponding to two types of dynamic loading. These durability tests are from two different types of PEMFCs, an open cathode/dead-end anode PEMFC (hereafter referred to as FC-1) and a vehicle-oriented commercial PEMFC (hereafter referred to as FC-2). The operating conditions of FC-1 and FC-2 are shown in Table 2.

Table 2 Ageing experiments operating conditions

| Parameter | Value | |
|---|---|---|
|  | FC-1 | FC-2 |
| Active surface (cm$^2$) | 33.63 | 25 |
| Pressure at hydrogen inlet (bar) | 1.35 | 1.1 |
| Pressure at air inlet (bar) | 1.013 (i.e., 1 atm) | 1.1 |
| Nominal output power (W) | 73.5 | 23.14 |
| Operating temperature (°C) | 29.6 to 51.7 (Corresponding to current) | 85 |
| Number of cells | 15 | 1 |
| Temperature regulate mode | 24-V dc air fan | External circulating water pump |
| Humidity regulate mode | Non-humidifier (self-humidified) | Built-in humidifier |

The FC-1 is designed for hybrid electric bicycles and its mission profile corresponds to the demands of real operating conditions. The stack of FC-1 contains 15 cells with an integrated 24V DC fan for air supply and temperature regulation. The operating temperature of FC-1 is related to the load current, which can be expressed by the following equation.

$$T_{FC1} = 2.5074 I_{FC1} + 30.3585 \tag{7}$$

where $T_{FC1}$ and $I_{FC1}$ are the operating temperature and load current of FC-1, respectively. In addition, FC-1 is self-humidifying and performs a purge lasting 0.5 s per 30 s [14]. The mean cell voltage profile from the dynamic load cycle of FC-1 is shown in Figure 4 (a), whereas the current profile is shown in Table 3 and Figure 4 (c). Each dynamic cycle lasts approximately 2.5 hours, in which the FC-1 load current is organized into seven test steps that switch repeatedly between 0 A and 8 A. In this way, the start-up/standby operating conditions in the hybrid system are simulated [16]. More details about FC-1 durability testing and ageing data can be found in our previous works [14,16].

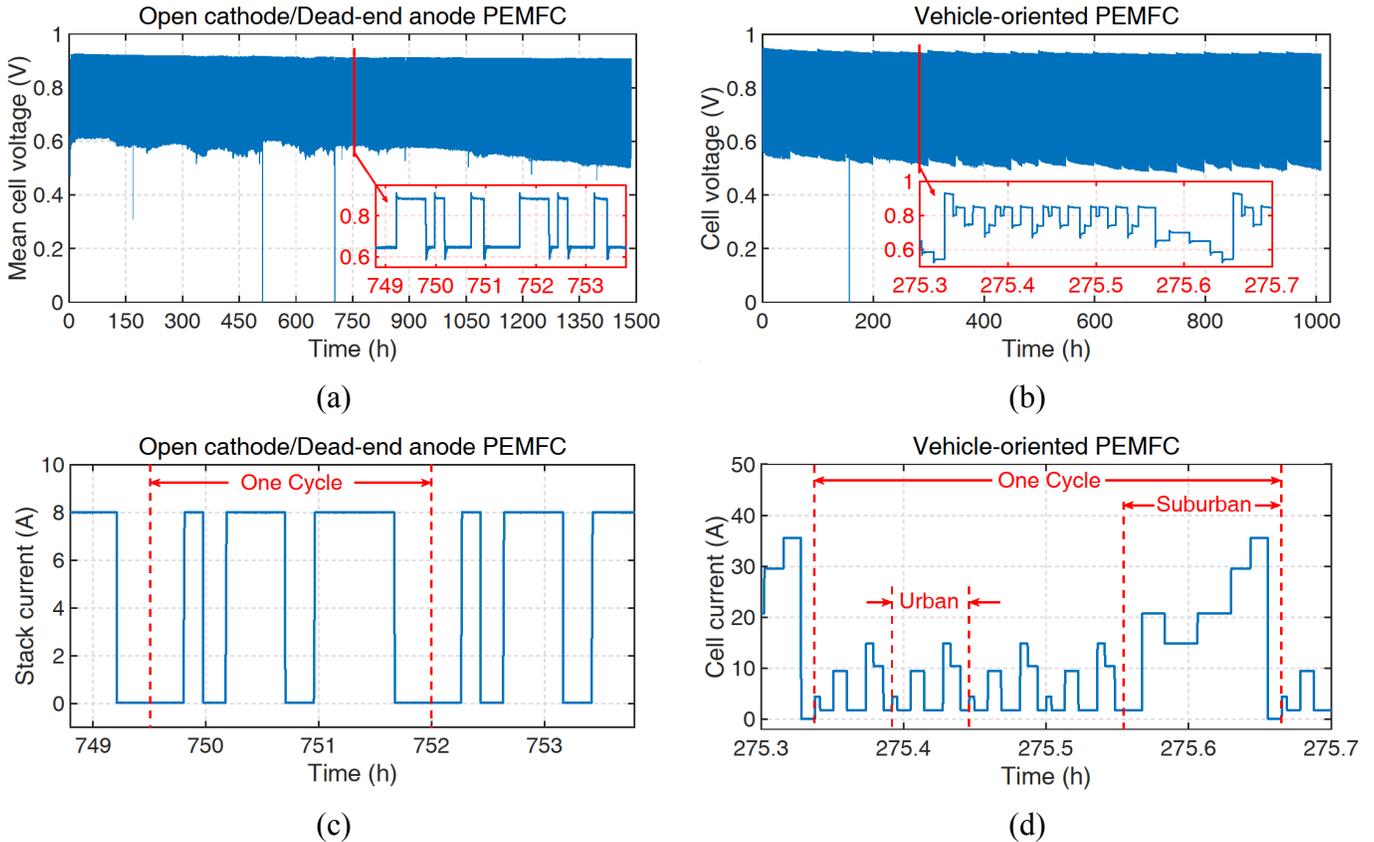

Figure 4. Dynamic load cycle profiles for fuel cell aging experiments, Voltages: (a). from FC-1, (b). from FC-2; Currents: (c). from FC-1, (d). from FC-2.

The FC-2, produced by Wuhan New Energy Co., Ltd, has a mission profile that corresponds to the demands of actual vehicle operating conditions. The stack of FC-2 contains a single cell. Ageing experiments are deployed on the Greenlight test station. The built-in humidifier is placed at the cathode and anode. The operating temperature is regulated using an external circulating water pump [20]. The cell voltage profile of the FC-2 dynamic load cycle is shown in Figure 4 (b), and the current profile is shown in Table 3 and Figure 4 (d). In this case, one dynamic cycle lasts approximately half an hour and involves two types of conditions, named "Urban" and "Suburban", respectively. Each of these conditions can be divided into seven test steps, as shown in Table 3. Specifically, the "Urban" condition is designed to simulate the frequent load switching of the vehicle at low speeds, while the "Suburban" condition is designed to simulate the vehicle switching between loads at medium/high speeds. In each dynamic cycle, "Urban" is repeated four times and then

"Suburban" is executed once [20]. More details about FC-2 and ageing data can be found in our previous works [20,33,44].

Table 3 Current profiles of dynamic load cyclic

| Test step | FC-1 | | FC-2 | | | |
| --- | --- | --- | --- | --- | --- | --- |
| | | | Urban condition | | Suburban condition | |
| | Duration (s) | Current (A) | Duration (s) | Current (A) | Duration (s) | Current (A) |
| 1 | 1060 | 0 | 13 | 4.45 | 46 | 1.78 |
| 2 | 630 | 8 | 33 | 1.78 | 58 | 20.75 |
| 3 | 710 | 0 | 35 | 9.51 | 82 | 14.85 |
| 4 | 1910 | 8 | 47 | 1.78 | 85 | 20.75 |
| 5 | 915 | 0 | 20 | 14.85 | 50 | 29.65 |
| 6 | 2565 | 8 | 25 | 10.4 | 44 | 35.6 |
| 7 | 1065 | 0 | 22 | 1.78 | 36 | 0 |

**Remarks**: (1) The load demands faced by PEMFC are complex and variable, which makes it steeply difficult to predict future stack voltages. The inherent degradation evolution is also masked by the dynamic mission profile, resulting in observations such as voltages that do not intuitively reflect long-term performance loss behavior. Therefore it is hard to achieve RUL estimation by directly predicting future voltages. (2) The PEMFC long-term durability tests are extremely sparse, and open-source datasets in particular are lacking.

## 5 Prognostics performance evaluation

In this work, the prognostics of fuel cells under two dynamic load conditions is investigated. The prognostics process is implemented based only on historical data of the fuel cell stack in question. Training and testing of prediction models can be achieved using data from a single long-term fuel cell test.

### 5.1 Evaluation metrics

Among various evaluation metrics, especially for RUL prediction performance, the prognostics horizon (PH) and the relative accuracy (RA) are considered both effective and widely used [1,11,12,16,17,19]. For the *i*-th failure threshold, a confidence region ($CR_i^{PH}$) is set. This region is calculated by the following equation.

$$RUL_i - \alpha_{low}EOL_i \leq CR_i^{PH} \leq RUL_i + \alpha_{up}EOL_i \tag{8}$$

where $RUL_i$ is the actual remaining useful life. $EOL_i$ is the actual end-of-life, which represents the operating time corresponding to the failure threshold. $\alpha_{low}$ and $\alpha_{up}$ represent the lower and upper bounds accuracy modifiers of $CR_i^{PH}$, respectively. The smaller the values of $\alpha_{up}$ and $\alpha_{low}$, the narrower the confidence region and the more stringent the evaluation criteria. In fact, the prediction results are expected to fall into a confidence region as narrow as possible, which means that the method has a high and reliable prediction accuracy. Moreover, $\alpha_{up}$ and $\alpha_{low}$ can be equal (e.g., in the literature [46]) or different (e.g., in the literature [47]). Commonly, if the predicted result is larger than the actual RUL, it means that the PEMFC will actually reach failure earlier than expected. In this condition, the PEMFC will potentially suffer a sudden failure or even a fatal shutdown. In contrast, when the predicted result is less than the actual RUL, the actual failure of the PEMFC lags behind the expectation. While this is again less accurate, this allows more ample time for operational control and/or maintenance decisions and is more acceptable. Based on the above factors, in this work, $\alpha_{up}$ and $\alpha_{low}$ are set to 0.1 and 0.2, respectively. This represents a slightly more generous tolerance for predicted results to be less than the actual RUL. Assuming that all the predicted RULs fall into $CR_i^{PH}$ after the prognostics point $t_i^{PH}$, the prognostics horizon ($PH_i$) at this point can be calculated as follows,

$$PH_i = EOL_i - t_i^{PH} \tag{9}$$

In addition, the confidence region is further defined stringently, for instance, using the $\alpha-\lambda$ performance to set the confidence region ($CR_i^{\alpha-\lambda}$) as follows,

$$RUL_i(1 - \alpha_{low}) \leq CR_i^{\alpha-\lambda} \leq RUL_i(1 + \alpha_{up}) \tag{10}$$

The confidence range $CR_i^{\alpha-\lambda}$ tightens over time, requiring more demanding prognostics performance.

Moreover, it is essential to calculate the relative accuracy (RA) of the predicted RUL for quantitatively evaluating the prediction performance. The calculation procedure is as follows,

$$RA_i = 1 - \frac{|RUL_i - R\widehat{U}L_i|}{RUL_i} \tag{11}$$

where the relative accuracy $RA_i$ corresponding to the $i$-th failure threshold. $R\widehat{U}L_i$ is the predicted remaining useful life.

### 5.2 Evaluate the feasibility of the extracted HI

In this work, a set of fluctuating signals and a residual component are separated from the stacked voltage signal under dynamic load using the HHT-based HI extraction method. Among them, the residual signal is a low frequency and null component that has the ability to characterize the physical features of the signal [24]. Moreover, the residual component can also reflect the trend characteristics of the original signal [31]. This allows to reveal the inherent degradation trend characteristics of the PEMFC, and therefore the residual component is chosen as HI. The feasibility of the extracted HI is evaluated in this subsection.

The programs used in this paper are developed in Python 3.7.11, Keras version 2.4.3, and Tensorflow version 1.15.0 backend software environments. They are deployed on a desktop computer containing an Intel Xeon E3-1230-v3 processor @ 3.3 GHz and 16 GB memory.

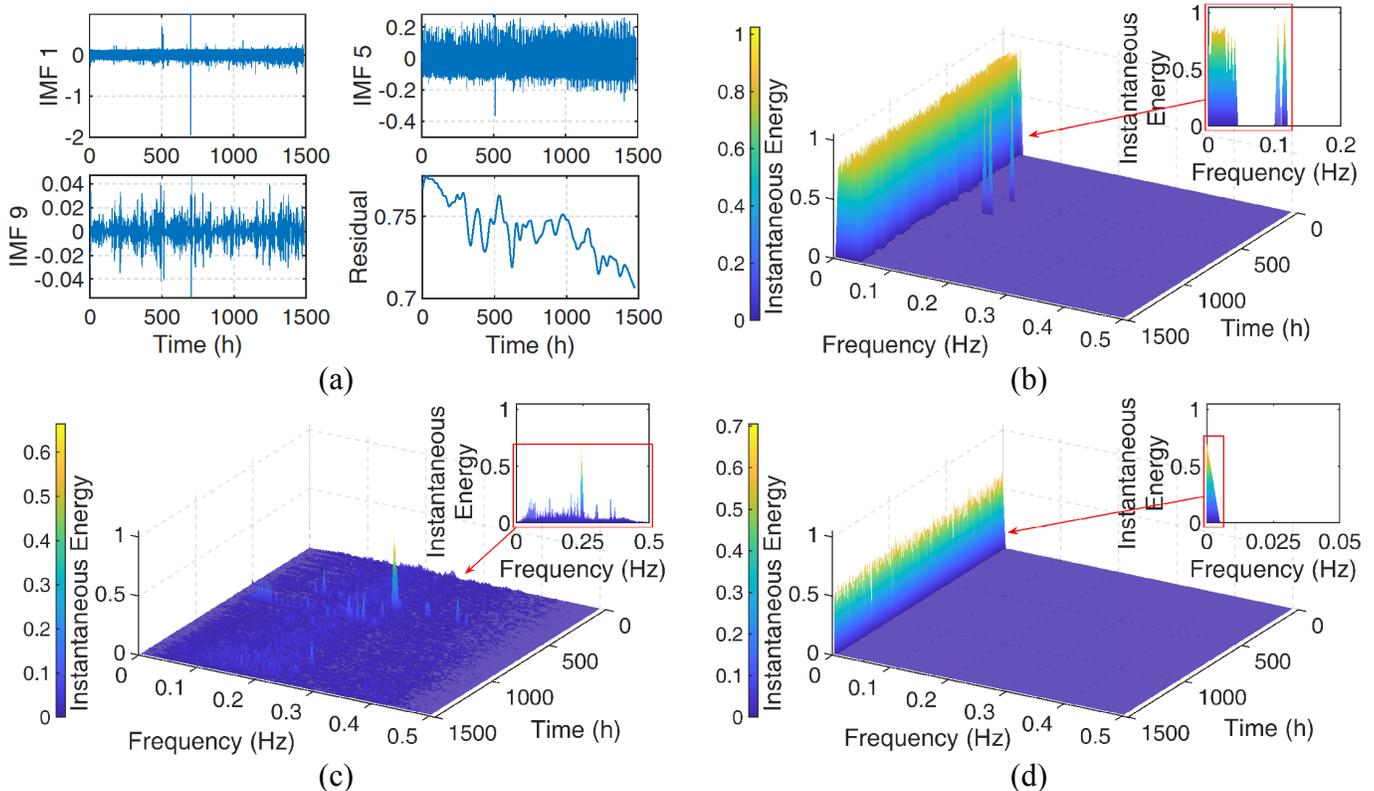

Figure 5. For FC-1: (a) IMFs and residual obtained by EMD; Time-frequency-energy spectrum: (b) the dynamic voltage, (c) all IMFs, (d) the residual.

The HHT-based method proposed in Section 2 is utilized to extract the HI, and the threshold of IF is set to 0.005 Hz. The dynamic voltage of FC-1 is decomposed into eleven IMFs and one residual, while the dynamic voltage of FC-2 is decomposed into nine IMFs and one residual. Figure 5 (a) and Figure 6 (a) show the partial (1st, 5th, 9th) IMFs and the final residual of FC-1 and FC-2, respectively. Besides, Figure 5 (b), (c) and (d) show the time-frequency-energy spectrum of the dynamic voltage (original signal), all IMFs, and residual of FC-1, respectively. Similarly, Figure 6 (b), (c) and (d) correspond to FC-2. The HHT-based HI extraction method square effectively separates the high-frequency features (IMFs) and the low-frequency feature (residual) of the dynamic voltage signal. By analyzing the instantaneous energy distribution of IMFs, it facilitates to identify fuel cell abnormal operation. For instance, in Figure 5 (c) and Figure 6 (c), most of the instantaneous energies are at low levels. Whereas several obvious instantaneous energy anomaly peaks can be found, which correspond to the abnormal operation of FC-1 and FC-2 mentioned in our previous work [14,16,20,44]. Therefore, the HHT-based extraction method not only has the ability to isolate dynamic operations in the stack voltage signal, but also to filter potential noise and fast phenomena. In contrast, the residual retains a relatively high level of instantaneous energy, as in Figure 5 (d) and Figure 6 (d). The instantaneous energy distribution of the residual shows a natural decreasing trend along the ageing time. This allows the residual to indicate the intrinsic degradation process of stack voltage.

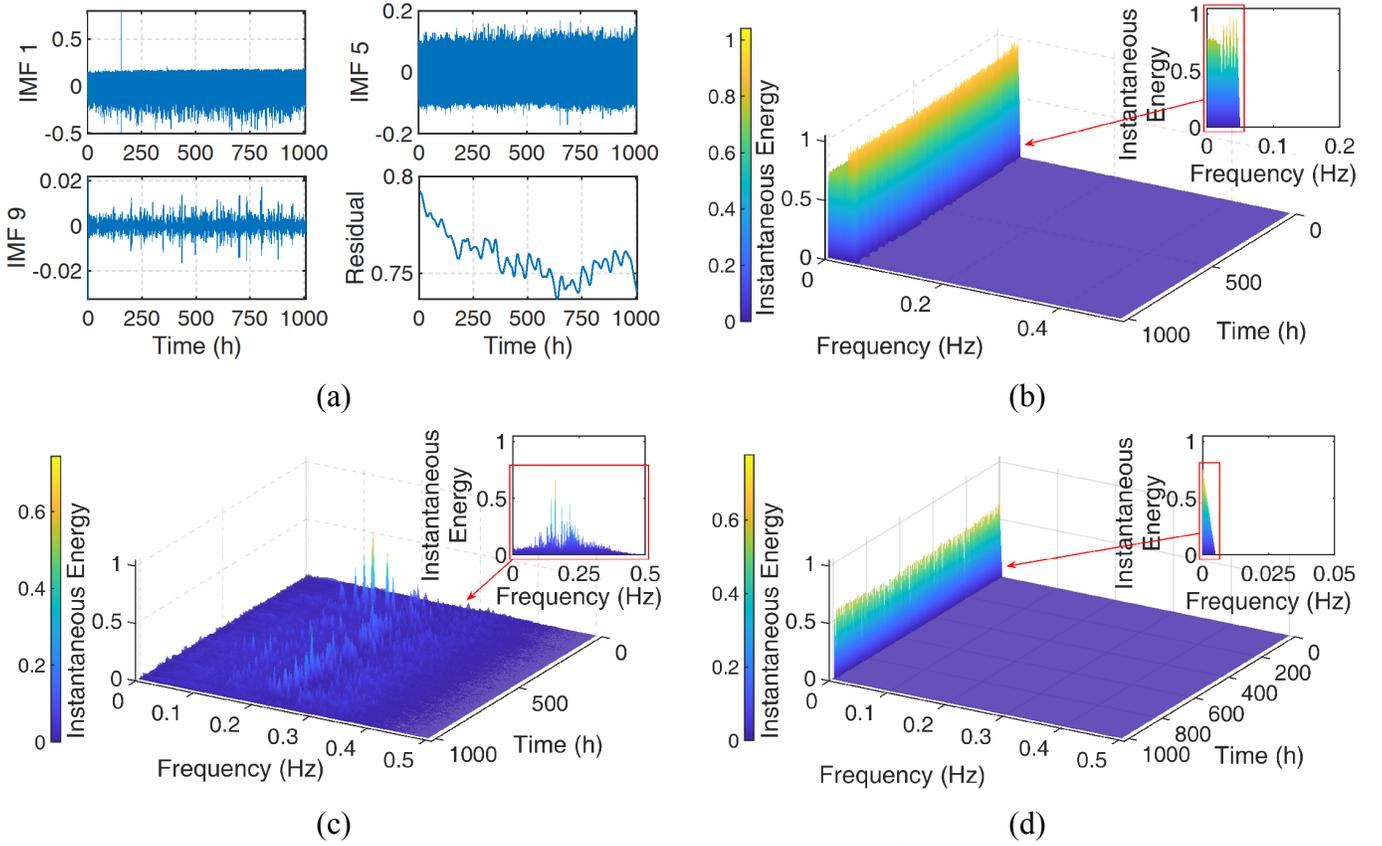

Figure 6. For FC-2, (a). IMFs and residual obtained by EMD; Time-frequency-energy spectrum: (b) the dynamic voltage, (c) all IMFs, (d) the residual.

In addition, the rationality of the proposed method is evaluated by comparing it with two other HI extraction methods. The first method uses the degradation model proposed in [15]. The equivalent internal resistance is obtained by piecewise linear regression and the virtual steady-state voltage is reconstructed as HI. Hereinafter, it is called the "Curve-fitting" method. The second method utilizes the linear parameter-varying (LPV) model structure, as in [14], called autoregressive model with exogenous input (ARX). The virtual steady-state voltage is extracted as HI by combining time-varying properties identification and sliding-window methods. Hereinafter, it is called "LPV-ARX" method. In the comparison experiments, the HIs extracted by the three methods are rescaled to the range of 0 to 1 and are processed using a uniform normalization as in the following equation.

$$\widetilde{HI}_t = \frac{HI_t - HI_{max}}{HI_{max} - HI_{min}} \tag{12}$$

where $HI_t$ is the extracted health indicator at operating time $t$. $HI_{max}$ and $HI_{min}$ are the maximum and minimum values in the sequence of health indicator, respectively. $\widetilde{HI}_t$ denotes the rescaled health indicator at time $t$.

As in Figure 7, all three methods match well on both PEMFC stacks, besides a few outlier points. In particular, they all capture the inherent voltage degradation trend. The proposed HHT-based HI extraction method possesses satisfactory feasibility. This is crucial for both analyzing the PEMFC degradation and predicting the RUL.

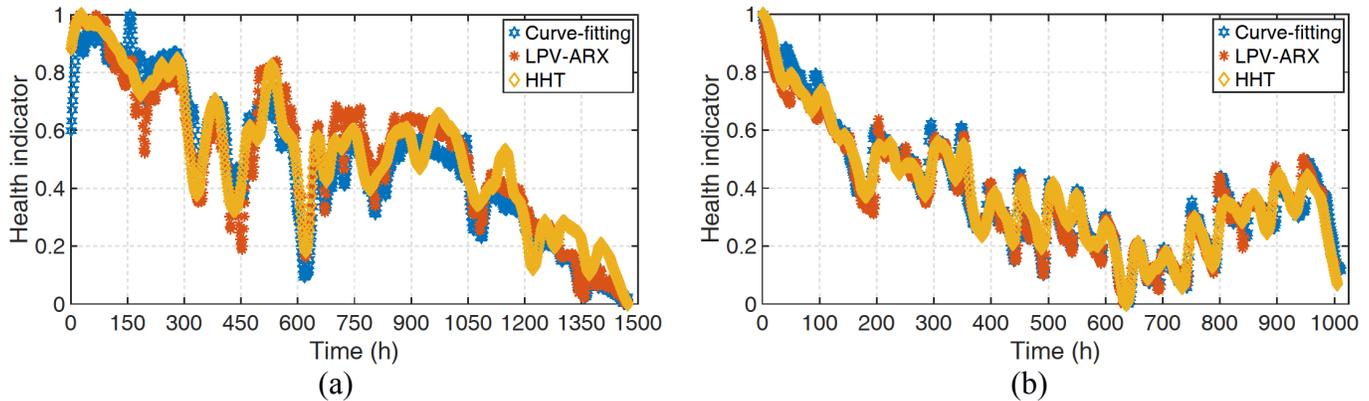

Figure 7. Comparison of health indicators extracted based on HHT with the other two methods, for (a) FC-1; (b) FC-2.

Further, the computational costs of the three methods are compared, as in Table 4. LPV-ARX has the highest computational cost and is significantly more than the other methods. The computational cost of HHT is slightly lower than that of Curve-fitting. It is noteworthy that the FC-1 ageing test duration is about 50% longer than the FC-2. The increase in ageing data causes the computational cost of LPV-ARX to expand significantly. In contrast, for the other two methods, the computational cost impact of the increased ageing data is not significant. Overall, the HHT-based approach is computationally inexpensive and has the potential to provide superior real-time performance.

Table 4 Computational cost comparison of the three methods

| Method | Execution time (s) | |
|---|---|---|
|  | FC-1 | FC-2 |
| Curve-fitting | 55.61 | 51.85 |
| LPV-ARX | 1167.77 | 518.42 |
| HHT | 21.73 | 16.93 |

**Remarks**: (1) The normalization of HI is for full life cycle/whole aging test, in which case the maximum and minimum values of HI are not time-varying; (2) The normalization of HI is to avoid the influence of different numerical intervals on the method evaluation and also to facilitate setting the same-level failure thresholds; (3) The normalization of HI is not mandatory, and the proposed prognostics is also valid for the original HI numerical interval.

## 5.3 Evaluate the predicted lifetime

For properly evaluating the RUL prediction performance, nine prognostics test points are set for FC-1, as in Figure 8 (a), with 147.5 hours between each two prognostics points. Meanwhile, twelve prognostics test points are set for FC-2, as in Figure 8 (b), with 49 hours between each two prognostics points. The United States Department of Energy defined end-of-life as a 10% loss of initial performance, which is suitable for constant current operating conditions [48]. However, when a PEMFC operates in dynamic load conditions, there is still no agreement on the definition of the FT [47]. In addition, selecting a fixed failure threshold seems to ignore the effect of uncertainty. In order to be able to more appropriately evaluate the prognostics performance and consistency under different failure thresholds, ten failure thresholds ranging from 0 to 0.09 spaced at 0.01 are set in this work.

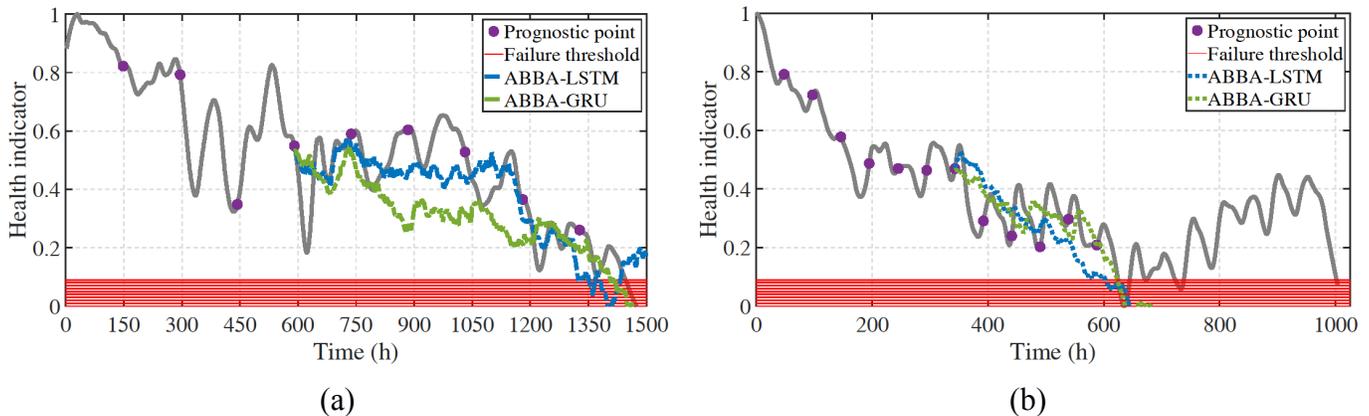

(a)  (b)
Figure 8. Prognostics points set on health indicators, for (a) FC-1, (b) FC-2; Comparison of trends predicted by ABBA-LSTM and ABBA-GRU, (a) from FC-1 at hour 590, (b) from FC-2 at hour 343.

Twenty ABBA-GRU models are deployed with random initialization of the weight matrix, and for each model, the configuration is as follows: ABBA tolerance is set to 0.001; The types of clusters coefficient $k$ is chosen adaptively between 1 and 100; Two GRU layers are included, each with 50 hidden units. The training process is optimized using Adam [45], the training learning rate is set to 0.001, the batch size is 128, and the maximum training epoch is 10,000. These hyperparameters are set in a trial-and-error way. On the one aspect, the parameters should be assigned to achieve satisfactory ABBA representation. On the other aspect, the precision of RUL estimation should also be regarded.

In order to concretize the enhancement of the GRU model on the prediction performance, a comparison experiment between the ABBA-LSTM model and the ABBA-GRU model is arranged. In this case, the ABBA-LSTM model is set up in the same way as [16]. For FC-1, one of the degradation trends predicted by each model at hour 590 is shown in Figure 8 (a). For FC-2, one of the degradation trends predicted by each model at hour 343 is shown in Figure 8 (b). The degradation trends predicted by the two models are similar, but there are differences in the computational costs. In comparison, the computational cost of the ABBA-GRU model is reduced by more than 30%, as in Table 5.

Table 5 Computational cost comparison of ABBA-LSTM and ABBA-GRU

| Method | Training duration (s) | | Prediction duration (s) | |
| --- | --- | --- | --- | --- |
| | FC-1 | FC-2 | FC-1 | FC-2 |
| ABBA-LSTM | 689.42 | 524.26 | 53.59 | 49.22 |
| ABBA-GRU | 465.72 | 354.15 | 37.28 | 34.24 |

In another comparison experiment, two state-of-the-art methods, the auto-regressive integrated moving average (ARIMA) method and the ESN method, are introduced to evaluate the prediction performance of RULs. The ARIMA method is set up as in the literature [49], while the ESN method configuration is taken from the literature [14]. The results of the comparison experiments from FC-1 when the failure threshold is equal to 0 are shown in Figure 9 (a) and Figure 9 (b), which demonstrate the predicted RULs and RAs from the three methods, respectively.

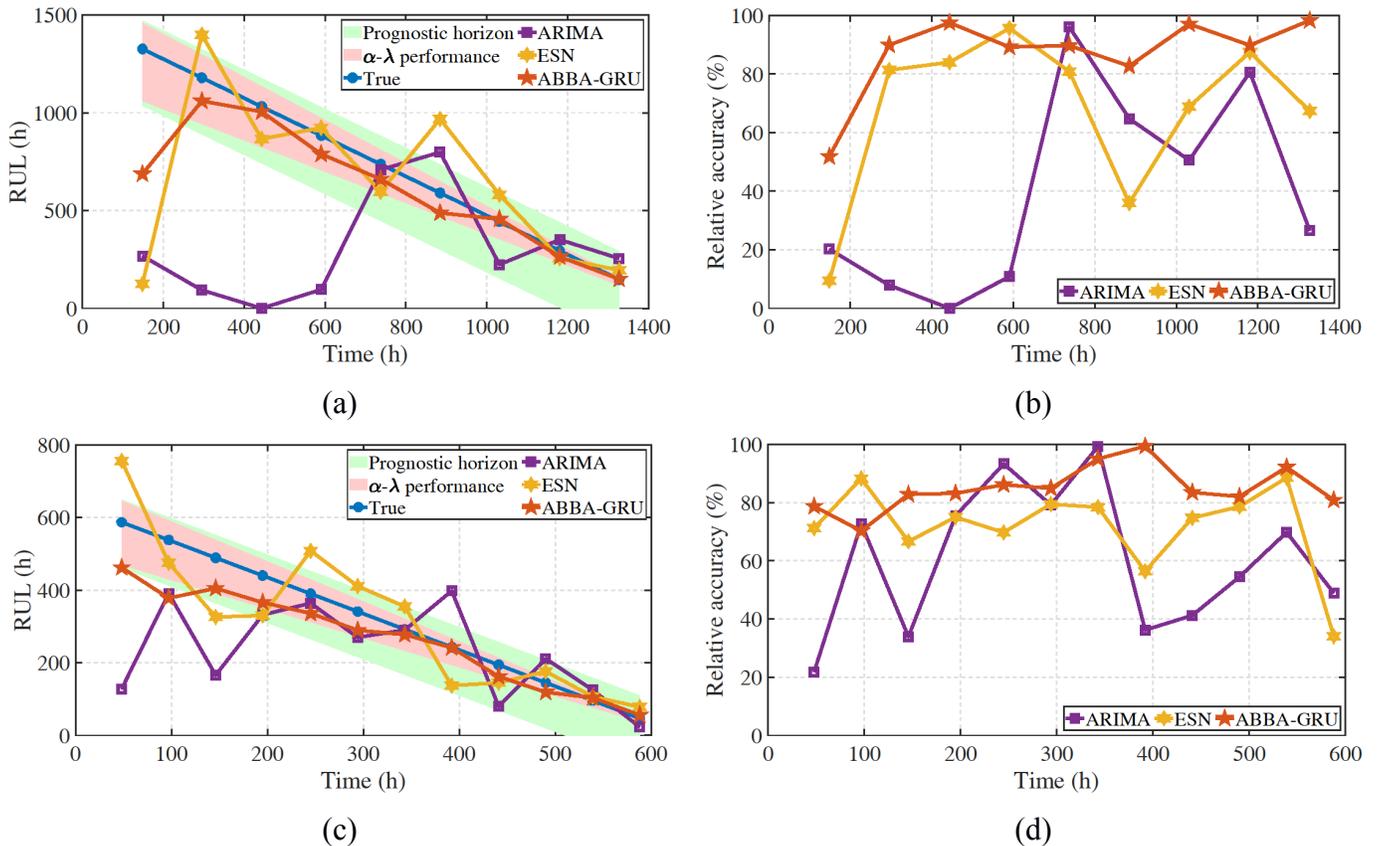

Figure 9. Prognostics comparison experiments with a failure threshold set to 0. (a) Predicted RULs for FC-1, (b) Relative accuracy for FC-1, (c) Predicted RULs for FC-2, (d) Relative accuracy for FC-2.

The ARIMA and ESN methods give the same PH of 296 hours, while ABBA-GRU provides a PH of about 1032 hours. Among the nine prognostics test points, ABBA-GRU meets the $\alpha$-$\lambda$ performance criteria eight times, ESN meets the criteria four times, and ARIMA meets the criteria once. In addition, the average RA of all the test points is about 87% for ABBA-GRU, 68% for ESN, and 40% for ARIMA. Similarly, Figure 9 (c) and Figure 9 (d) correspond to FC-2. The results show that ARIMA provides 49 hours of PH, while ESN provides 196 hours and ABBA-GRU provides 441 hours of PH. For $\alpha$-$\lambda$ performance, ABBA-GRU meets eight of the eleven test points, ESN meets one and ARIMA meets two. The average RA for the three methods is about 85% for ABBA-GRU, 72% for ESN, and 61% for ARIMA, respectively.

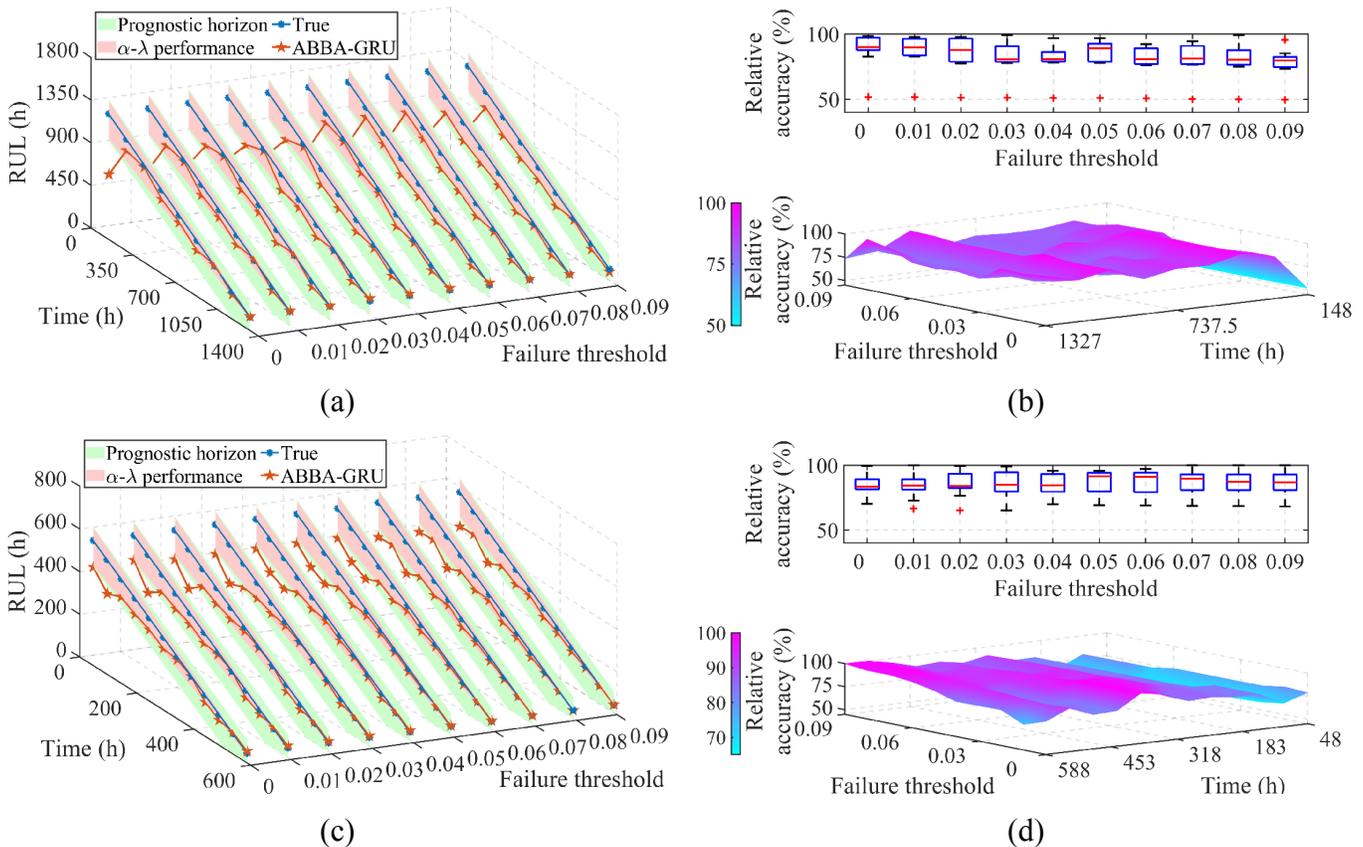

Figure 10. FC-1 at different failure thresholds: (a) predicted RULs, (b) distribution of RAs; FC-2 at different failure thresholds: (c) predicted RULs, (d) distribution of RAs.

The ABBA-GRU model is deployed at each prognostics point and the RUL is predicted. The RUL prediction results for FC-1 and FC-2 are shown in Figure 10 (a) and (c), respectively. At the initial prediction point, the prognostics performance is not optimal due to the lack of historical data. However, as the measurements continue to be acquired, the predicted results gradually converge to the actual RUL. For FC-1, all PHs reach about 1032 hours. Among all the 90 prognostics points under different failure thresholds, 63 points meet the $\alpha$-$\lambda$ performance requirements, with a passing rate of 70%. For FC-2, the maximum PH is 441 hours, the minimum is 392 hours, and the average is 411.6 hours. Among all the 120 prognostics points under different failure thresholds, 92 points met the $\alpha-\lambda$ performance requirements, with a passing rate of 77%.

Figure 10 (b) and (d) show the relative accuracies for FC-1 and FC-2, respectively. The red "+" in the upper part of Figure 10 (b) and (d) corresponds to the RA of outlier points. Overall, the accuracy of the prediction improves over time. The distribution of RA under different failure thresholds fluctuates slightly and is consistent overall. In summary, the prognostics performance of the proposed method is stable with satisfactory accuracy under different failure thresholds.

## 6  Conclusion

In this paper, a data-driven prognostics approach is proposed for fuel cells under dynamic operating conditions. A Hilbert-Huang transform-based method is utilized to extract the health indicator from the dynamic voltage of fuel cells. The historical health indicator data is used to train the ABBA-GRU model, which in turn predicts the degradation trends and the remaining useful life of the stack. Dynamic load ageing experiments are carried out on two different types of fuel cells, and the prognostics approach is evaluated with the ageing data. The following conclusions can be summarized from the experimental and simulation results.

- The extracted health indicators characterize the inherent degradation behaviour of the stack voltage and the extraction method is computationally low-cost.
- The symbolic-based gated recurrent unit model provides a credible prognostics horizon of up to 2-6 weeks.
- While obtaining similar prediction performance as the ABBA-LSTM model, the computational cost of the ABBA-GRU model is reduced by 30%.
- The ABBA-GRU model provides a competitive prognostics horizon and relative accuracy compared to the other two state-of-the-art methods.
- The ABBA-GRU model exhibits satisfactory relative accuracy and consistency under multiple failure thresholds.

The above approach enhances real-time performance while providing a credible prognostics horizon and appropriate prediction accuracy. It facilitates the online deployment of the prognostics method. In future work, efforts will be made to explore prognostics methods incorporating fault diagnostics, e.g., adaptively setting failure thresholds by analyzing abnormal behavior. Meanwhile, prognostics-based operational control decisions will be investigated to extend the fuel cell lifetime.

## Acknowledgement

This work was supported in part by the China Scholarship Council (CSC) under Grant [grant number 201906290107].

## Appendix A

For the $i$-th residual $r_i(t)$, where $i = 1, \ldots, n$. The Hilbert transform of the residual (e.g., $\mathcal{H}_i(t)$) can be calculated as follow,

$$\mathcal{H}_i(t) = \frac{1}{\pi} PV \int_{-\infty}^{\infty} \frac{r_i(\tau)}{t-\tau} d\tau \tag{A.1}$$

where $PV$ is the Cauchy principal value. The $z_i(t)$, which is an analytical function of $r_i(t)$, can be constructed as the following equation,

$$z_i(t) = r_i(t) + j\mathcal{H}_i(t) = a_i(t)e^{j\phi_i(t)} \tag{A.2}$$

where $j$ is the imaginary unit. The amplitude function $a_i(t)$ and the instantaneous phase function $\phi_i(t)$ can be expanded as following equation,

$$\begin{cases} a_i(t) = \sqrt{r_i^2(t) + \mathcal{H}_i^2(t)} \\ \phi_i(t) = \arctan[\mathcal{H}_i(t)/r_i(t)] \end{cases} \tag{A.3}$$

The instantaneous frequency $\omega_i(t)$ is calculated by the following equation,

$$\omega_i(t) = \frac{1}{2\pi} \frac{d\phi_i(t)}{dt} \tag{A.4}$$

Then, the instantaneous energy $\varepsilon_i(t)$, corresponding to $r_i(t)$, is calculated by the following equation,

$$\begin{cases} \varepsilon_i(t) = \int_{\omega_i^-}^{\omega_i^+} H^2(\omega_i, t) d\omega_i \\ H(\omega_i, t) = Re[a_i(t)e^{j\phi_i(t)}] \end{cases} \tag{A.5}$$

where $\omega_i^-$ and $\omega_i^+$ are the frequency range upper and lower bounds of $r_i(t)$. The $H(\omega_i, t)$ is the Hilbert spectrum of $r_i(t)$, and $Re[f(x)]$ denotes to extract the real component of the function $f(x)$.

# Appendix B

Each recurrent unit of the GRU has the ability to adaptively capture the dependencies at different time scales. Similar to the LSTM unit, the GRU has gating units that regulate the flow of information inside the unit, but there is no separate memory unit [37].

The reset gate is used to modulate the extent to which the unit resets its activation or content. In this case, the reset can also be considered as the reading of the first symbol in the sequence. The reset gate ($r_t$) can be calculated by the following equation,

$$r_t = \sigma(W_r a_t + U_r h_{t-1} + b_r) \tag{B.1}$$

where σ represents the sigmoid activation function. $a_t$ is the current symbol, and $h_{t-1}$ is the previous hidden state. Further, the update gate ($z_t$) can be obtained with a similar equation,

$$z_t = \sigma(W_z a_t + U_z h_{t-1} + b_z) \tag{B.2}$$

Combined with the modulation of the reset gate, a candidate activation, also known as the unit internal hidden state ($\tilde{h}_t$), can be computed as follow,

$$\tilde{h}_t = \tanh[W_h a_t + U_h(r_t \odot h_{t-1}) + b_h] \tag{B.3}$$

where tanh represents the hyperbolic tangent activation function. The operator "$\odot$" denotes Hadamard product. In addition, in Equations (B.1)-(B.3), $W_x$, $U_x$, and $b_x$ are the input weight matrix, the unit internal weight matrix, and the bias vector, respectively. Among them, the subscript $x$ corresponds to $r$ (reset gate), $z$ (update gate), and $h$ (implicit state).

Subsequently, the current hidden state ($h_t$) is output, as in the following equation.

$$h_t = (1 - z_t) \odot h_{t-1} + z_t \odot \tilde{h}_t \tag{B.4}$$

It is worth noting that the current hidden state is also the predicted symbol, i.e., the notation "$\hat{a}_t$" in this paper.